\documentclass[letterpaper, 10 pt, conference]{ieeeconf}  

\IEEEoverridecommandlockouts                              

\usepackage{amsmath} 
\usepackage{amssymb}  
\usepackage{bm}
\usepackage{graphicx}
\usepackage{subfigure}
\usepackage{float}
\usepackage{multirow}
\usepackage{booktabs}
\usepackage{cite}
\usepackage{textcomp}
\usepackage[linesnumbered,ruled]{algorithm2e}

\providecommand{\norm}[1]{\left\lVert#1\right\rVert}
\usepackage{makecell}
\makeatletter
\let\NAT@parse\undefined
\makeatother
\usepackage{hyperref}

\hypersetup{
	colorlinks,
	citecolor=blue,
}

\title{\LARGE \bf ViLiVO: Virtual LiDAR-Visual Odometry for an Autonomous Vehicle with a Multi-Camera System}

\author{Zhenzhen Xiang, Jingrui Yu, Jie Li and Jianbo Su
\thanks{This work was partially financially supported by the projects of National Natural Science Foundation of China under grant 61533012 and 91748120.}
\thanks{Zhenzhen Xiang, Jingrui Yu and Jianbo Su are with the Department of Automation, Shanghai Jiao Tong University, Shanghai 200240, China. (e-mail: zzxiang.sjtu@gmail.com, \{yujingrui, jbsu\}@sjtu.edu.cn).
}
\thanks{Jie Li is with SAIC Motor, Shanghai, China (e-mail: lijie06@saicmotor.com).}
}

\begin{document}

\maketitle
\thispagestyle{empty}
\pagestyle{empty}

\begin{abstract}

In this paper, we present a multi-camera visual odometry (VO) system for an autonomous vehicle. Our system mainly consists of a virtual LiDAR and a pose tracker. We use a perspective transformation method to synthesize a surround-view image from undistorted fisheye camera images. With a semantic segmentation model, the free space can be extracted. The scans of the virtual LiDAR are generated by discretizing the contours of the free space. As for the pose tracker, we propose a visual odometry system fusing both the feature matching and the virtual LiDAR scan matching results. Only those feature points located in the free space area are utilized to ensure the 2D-2D matching for pose estimation. Furthermore, bundle adjustment (BA) is performed to minimize the feature points reprojection error and scan matching error. We apply our system to an autonomous vehicle equipped with four fisheye cameras. The testing scenarios include an outdoor parking lot as well as an indoor garage. Experimental results demonstrate that our system achieves a more robust and accurate performance comparing with a fisheye camera based monocular visual odometry system.

\end{abstract}

\section{Introduction}

\begin{figure}[t]
  \centering
  \includegraphics[width=2.8in]{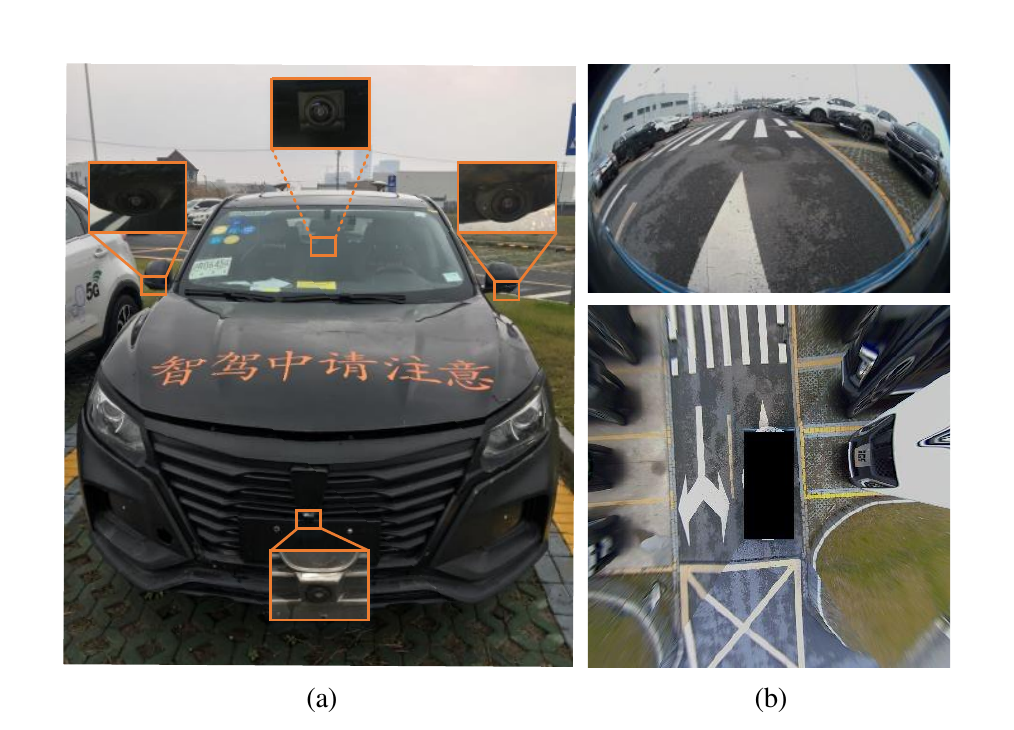}
  \caption{(a) Our testing autonomous vehicle with four surround-view fisheye cameras. (b) A sample image of the front camera (top) and a synthesized surround-view image after perspective transformation (bottom).}
  \label{introduction}
\end{figure}

For an autonomous vehicle, localizing its pose according to the landmarks in the environment is considered as one of its basic functionalities. Recently, localization with visual sensors such as cameras has attracted a lot of interest, which is often called visual odometry (VO), or visual simultaneous localization and mapping (VSLAM) for a more comprehensive system. Although it is still a long way to design a universal VO system suitable for all applications, huge progress has been achieved on deploying the system to some specific scenarios, for example, to provide an autonomous parking service in a parking lot\cite{schwesinger2016automated}. To acquire an accurate and robust performance, the vision system mounted a on vehicle usually consists of multiple high-resolution cameras with different field of view (FoV). A typical image with a wide FoV can be captured by a fisheye camera, which has been widely used for parking assistance, as shown in Fig.~\ref{introduction}. A wider view means that more landmarks can be seen by the vehicle, which can directly improve the reliability of VO systems if the information is appropriately exploited.

Several approaches have been proposed to design a VO system for an autonomous vehicle with multiple cameras. Lee et al. \cite{hee2013motion} proposed a method to solve the problem of motion estimation with 2D-2D image point correspondences and pose estimation with 2D image point and 3D world point correspondences. Kneip et al. \cite{kneip2014efficient} presented a low-dimension and efficient algorithm to compute the relative pose of a multi-camera system for more general configurations, while Ventura et al. \cite{ventura2015efficient} further simplified the pose estimation problem with a first-order approximation for real-time applications. In \cite{forster2016svo}, Forster et al. extended their previous Semi-direct VO (SVO) to a multi-camera system with a benefit of the performance. Liu et al. \cite{liu2018towards} presented a robust VO algorithm for an arbitrary number of stereo cameras with near-infrared illumination which can enhance the performance of the system at night. In this work we simplify the problem to a 2D motion estimation task with a surround-view image shown in Fig.~\ref{introduction}(b) which reserves most of the valuable visual landmarks.

\begin{figure*}[!thb]
  \centering
  \includegraphics[width=6.5in]{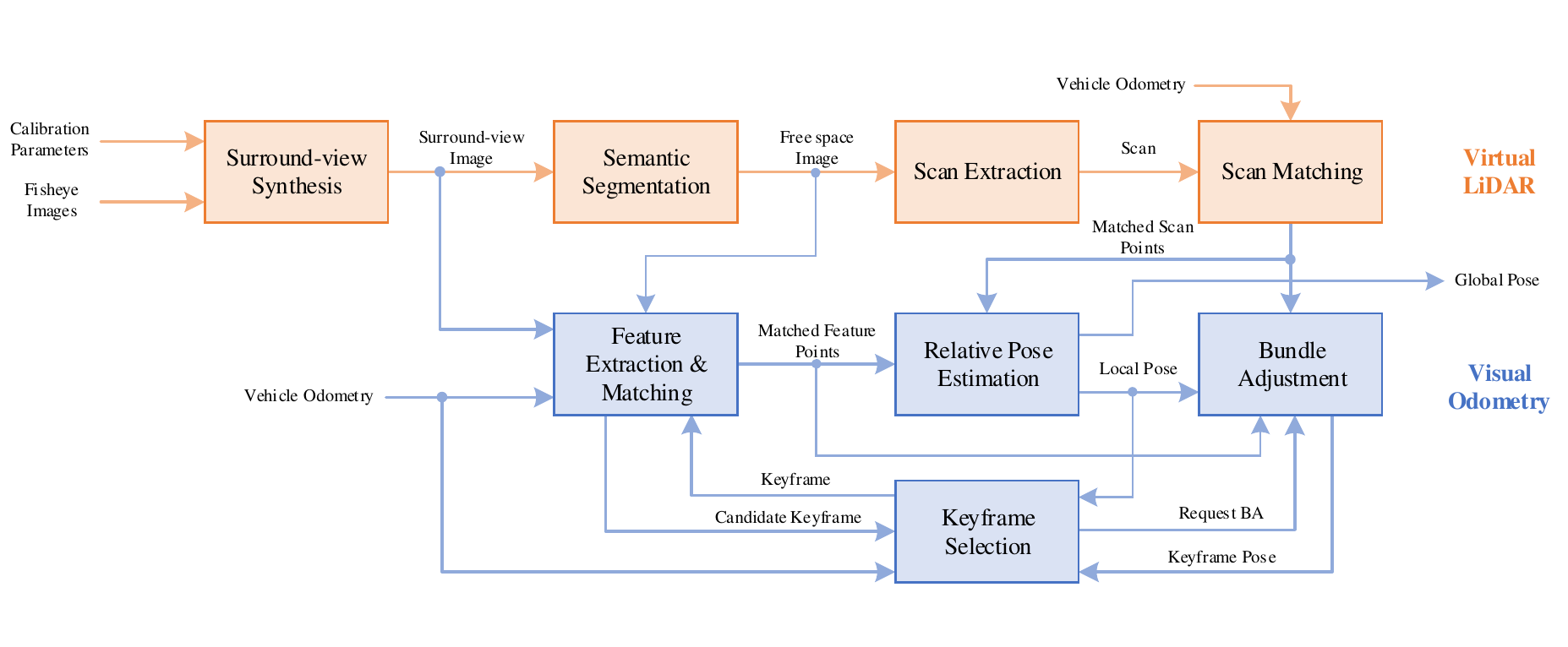}
  \vspace{-10pt}
  \caption{The scheme of the ViLiVO system. The scan matching results provided by the virtual LiDAR are fused and optimized with the feature matching results to construct a more robust visual odometry system.}
  \label{scheme}
\end{figure*}

With the rapid development of deep learning, more and more methods based on deep neural networks have been integrated into current VO or SLAM systems to boost the visual perception. Deep semantic segmentation model is one of the most widely used models. It can abstract higher level representations with pixel-wise accuracy, which improves the understanding of the environments. Li et al. \cite{li2018stereo} proposed a semantic method for both ego-motion estimation and 3D object tracking with the extracted contours and masks of the vehicles. In \cite{dube2018segmap}, a segment based map representation was proposed for 3D point clouds with extracted semantic information for localization only against static objects. To increase the robustness in dynamic environments, semantic masks were used to distinguish the static feature points and remove the outliers on moving objects \cite{yu2018ds}. Our proposed VO system also benefits from an extra processing with semantic segmentation results.

Although the stereo or multi-camera based VO can estimate the depth of feature points or pixels comparing with the monocular VO, more accurate measurements of the depth is more valuable. Light detection and ranging (LiDAR) sensors have been widely used for their high accuracy in measuring distance. Zhang et al. \cite{zhang2015visual} successfully combined the monocular camera with a 3D LiDAR for mapping and localization, which achieved low-drift and robust performance especially with aggressive movement. Graeter et al. \cite{graeter2018limo} proposed a LiDAR-monocular visual  odometry (LIMO) system further performing a foreground segmentation and plane fitting for the depth estimation of carefully selected landmarks. Lowe et al. \cite{lowe2018complementary} designed a handheld SLAM system simultaneously optimizing the measurements of a camera, a LiDAR and an IMU for a complementary perception purpose in both indoor and outdoor applications. Without using a practical LiDAR, we generate virtual 2D LiDAR scan data from the output of a semantic segmentation model which is much more economical.

In this paper, we aim to design a VO system for an autonomous vehicle with multiple fisheye cameras. Unlike previous work, we transform the original fisheye images to a surround-view image as the input of system. With such a transformation we can further extract the free space area around the vehicle from semantic segmentation results. Contours of the free space are simulated as a 2D LiDAR scan which is the output of our virtual LiDAR. Then we fuse the scan matching results with a feature-based VO to construct a novel virtual LiDAR-visual odometry (ViLiVO) system. Our system mainly combines the following advantages:

\begin{itemize}
  \item The synthesized surround-view image persists major information of original fisheye images since a large part of fisheye image is the ground as shown in Fig.~\ref{introduction}(b). The VO system still can benefit from the 360 degree view. Although objects with a certain height on the ground can be distorted, the features on the ground are well-undistorted and more suitable for tracking. With the calibration for perspective transformation, the scale between the image and the real world is derived, which directly eliminates the gap between 2D pixels on the surround-view image and 3D points on the ground.
  \item The semantic segmentation results as the free space area around the vehicle provide two kinds of information: a masked ground area and distance measurements to obstacles. The former can help to reject feature points on distorted objects, while the latter gives geometric information of the objects which is helpful when only few visual landmarks can be detected.
  \item Both the virtual LiDAR scan points and the visual landmarks are matched with a scan matcher and feature matcher, respectively. All of them are optimized together to generate an estimation of the relative pose as well as the keyframe pose after a local BA, which constructs a more robust VO than pure scan matching based or feature matching based odometry.
\end{itemize}

\section{System Overview}
The scheme of our ViLiVO system is illustrated in Fig.~\ref{scheme}. For the part of virtual LiDAR, we first undistort all images from the fisheye cameras and warp to generate the surround-view image. Then the surround-view image is processed by a semantic segmentation model to detect the free space area in the image. Contours of the free space area are discretized to generate scan points. The scan point correspondences after scan matching and the free space mask image are both sent to VO as its input.

For the part of VO, feature points are detected in the masked surround-view image. A direct matching is performed with the constraints of vehicle kinematics model to get feature correspondences. To increase the robustness of the system, we use a distance threshold and the Random Sample Consensus (RANSAC) method to remove outliers. In addition, the pose of the newly added keyframe is optimized with the bundle adjustment method. The reprojection error of feature points and the scan matching error are involved in the optimization.

\section{Virtual LiDAR}
\begin{figure}[t]
  \centering
  \includegraphics[width=2.8in]{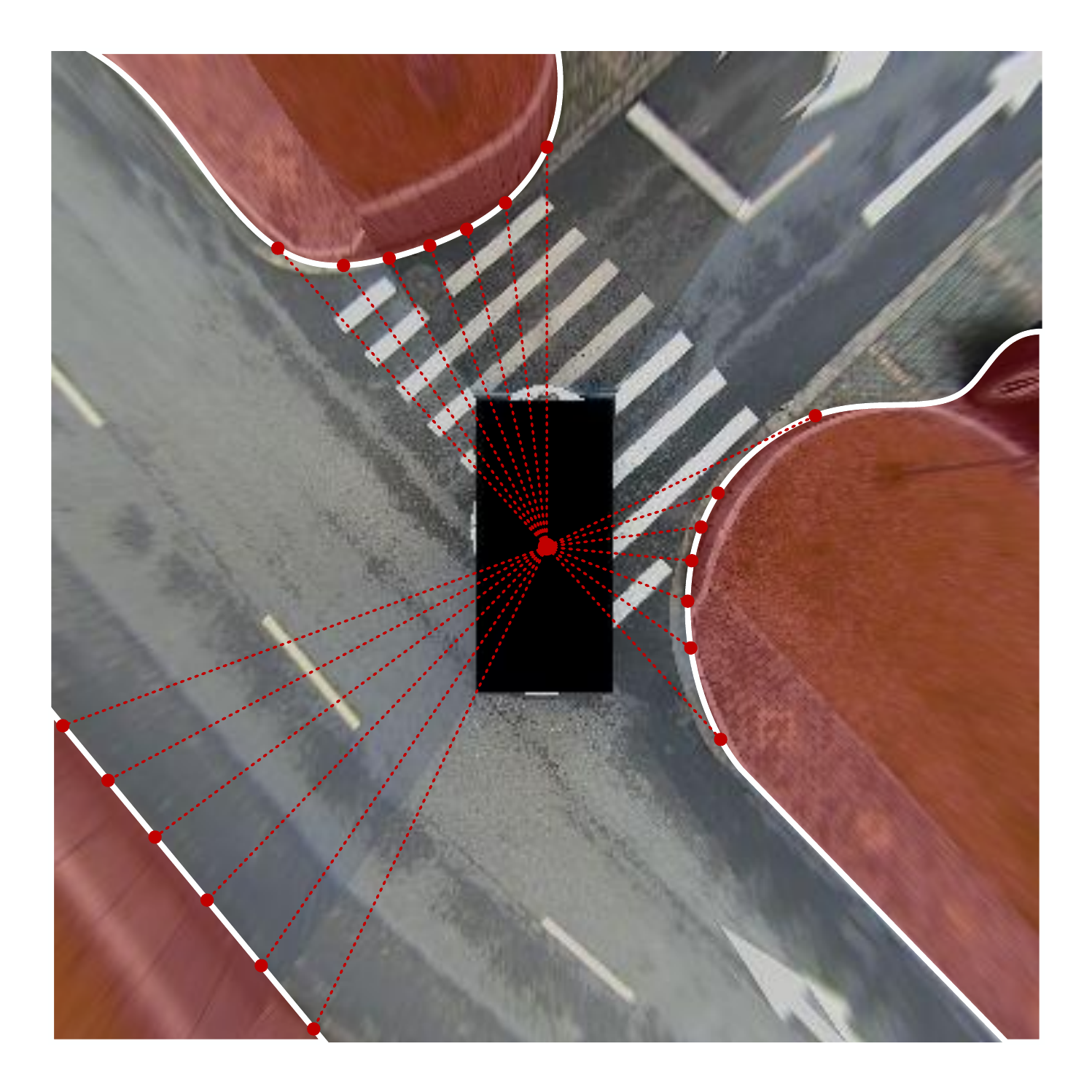}
  \caption{Demonstration of the proposed virtual LiDAR. The pixels in red are the segmented non-free space area. The virtual scan points are extracted on the contours of free space.}
  \label{virtual_lidar}
\end{figure}

A LiDAR sensor is able to measure the distance between objects in the environment and itself. The idea of the proposed virtual LiDAR directly comes from the semantic segmentation results of a surround-view image. The principle of our virtual LiDAR is shown in Fig.~\ref{virtual_lidar}.

Since major parts of fisheye camera images are the ground or objects on the ground, a surround-view image synthesized by a perspective transformation method can preserve most visual information. A deep neural network for semantic segmentation purpose is trained with free space and obstacles labeled to generate a free space mask image. Then we can simply extract the contours of the free space after applying morphological operations to the mask image, which can fill small holes and remove small separated parts.

Moreover, with the calibration in the perspective transformation progress, we can directly get the scale from pixel to meter. Therefore each point on the contours can be transformed to the vehicle frame with a practical scale. We combine all of the contour points and group the points with a certain angle step. Those points with the minimum distance in each angle window are used to simulate the final scan data.

\section{Visual Odometry}
Virtual LiDAR scans are beneficial to both motion estimation and obstacle detection. In this paper, we focus on the former and merge it with a feature based visual odometry system to increase its robustness and accuracy.

\subsection{Semantic Segmentation for Region of Interest}
To accurately calculate the relative pose between two surround-view images, first we need to ensure that all the detected feature points are on the ground. Those objects with a certain height above the ground will distort dramatically after the perspective transformation.

Fortunately, our virtual LiDAR can provide a mask to the surround-view image, which allows us to set the region of interest (ROI) to the free space area in the feature extraction module. All of those feature points outside the ROI are rejected.

\subsection{Relative Pose Estimation}
A keyframe based method is applied in our VO system. The relative pose of current frame to keyframe is calculated for a stable estimation as well as a lower drift.

\textbf{Motion prediction model:} The relative vehicle pose between two frames are calculated according to the Ackermann motion model. Detailed mathematical description of the model can be found in \cite{siegwart2011introduction}.

\textbf{Direct feature matcher:} The feature points in current frame can be projected to the keyframe with the relative pose transformation. Those points within a certain distance to the projected points are considered as potentially matched points. The initial feature correspondences are filtered later in outlier removal step.

\textbf{Scan matcher:} We use a 2D laser scan matcher proposed in \cite{censi2008icp} to perform the scan matching with an ICP (Iterative Closest Point) method and a point-to-line metric. Both the pose variation and the correspondences of scan points between current frame and keyframe are established.

\textbf{Outlier removal:} Since the estimation of the pose and landmarks is very sensitive to outliers, we apply a standard RANSAC method to filter both feature and scan point correspondences.

\textbf{Relative pose estimation:} With the filtered feature points matching and scan matching results, we can solve a least square problem to estimate the current pose with respect to the keyframe. Since we suppose all feature points are on the ground, the fix transformation from 2D image frame to 3D vehicle frame can be derived. The problem is formulated as the follows:
    \begin{equation}\label{eq_1}
    \begin{aligned}
        \underset{{\rm \mathbf{T}}^{K\!F}_{F_k}}{\rm argmin}\sum_iw_1\rho_\phi\left(\norm{\phi(p_{i,k},{\rm \mathbf{T}}^{K\!F}_{F_k})}^2_2\right)+\\
        \sum_jw_2\rho_\psi\left(\norm{\psi(q_{j,k},{\rm \mathbf{T}}^{K\!F}_{F_k})}^2_2\right),
    \end{aligned}
    \end{equation}
where ${\rm \mathbf{T}}^{K\!F}_{F_k}$ means the transformation of vehicle base from frame $k$ to current keyframe. The first and the second term of \eqref{eq_1} are feature points and scan points reprojection error with weights $w_1, w_2$ and loss functions $\rho_\phi(\cdot), \rho_\psi(\cdot)$. Here we choose the Cauchy function as the loss function to reduce the influence of noise. The feature point reprojection error
    \begin{equation}\label{eq_2}
        \phi(p_i,{\rm \mathbf{T}}^{K\!F}_{F_k})=\bar p_i - {\rm \mathbf{T}}^{K\!F}_{F_k} p_i
    \end{equation}
and the scan point reprojection error
    \begin{equation}\label{eq_3}
        \psi(q_j,{\rm \mathbf{T}}^{K\!F}_{F_k})=\bar q_j - {\rm \mathbf{T}}^{K\!F}_{F_k} q_j
    \end{equation}
are calculated for each feature point $p_{i,k}$ with its correspondence $\bar p_{i,k}$ in keyframe and each scan point $q_{j,k}$ with $\bar q_{j,k}$. It should be noticed that all of the feature points and scan points are transformed to the vehicle base frame before calculating the residual.

\subsection{Keyframe Optimization}
Keyframe is very important to both the short-term and long-term odometry, since all pose errors of keyframes will be accumulated and result in a drift. To achieve a better pose estimation for a keyframe, we take a certain number of frames and optimize them together with measurements of both feature points and scan points.

\textbf{Keyframe selection:} We consider both spatial and temporal variations to update the keyframe. For the spatial changes, first we check the pose estimation result given by the feature and scan matching process. If the deviation of translation or rotation exceeds corresponding thresholds, current keyframe will be updated by current frame after the optimization. Sometimes the cameras may fail to extract enough feature points or valid scans due to lack of texture or objects in the environment. So we also check the vehicle odometry data in such situations and decide whether the vehicle has moved a certain distance over the threshold.

For the temporal changes, we use the timestamp of each frame to check if the current keyframe has existed for too much time. With time updating, our VO system can quickly adapt to the changes of surroundings or light conditions.

\textbf{Bundle adjustment:} We take into account all of the frames between the current keyframe and the next candidate keyframe in BA. The poses for each frame and the feature points in current keyframe are optimized. The scan points are only used for residual calculation since the quality of scan points may decrease a lot with the movement of vehicle. Thus the BA problem can be formulated as the follows:
    \begin{equation}\label{eq_4}
    \begin{aligned}
        \underset{{\rm \mathbf{T}}^{K\!F}_{F_k}\in \mathcal{T},\bar p_{i,\!K\!F}\in \mathcal{P}}{\rm argmin} \sum_{i,k}w_1\rho_\phi\left(\norm{\phi(p_{i,k},{\rm \mathbf{T}}^{K\!F}_{F_k})}^2_2\right)+\\
        \sum_{j,k}w_2\rho_\psi\left(\norm{\psi(q_{j,k},{\rm \mathbf{T}}^{K\!F}_{F_k})}^2_2\right),
    \end{aligned}
    \end{equation}
where $\mathcal{T}$ is the set of frame poses and $\mathcal{P}$ is the set of feature points in current keyframe. Other notations are the same as \eqref{eq_1}.

\section{Experiments}
\subsection{System Setup}
\textbf{Hardware:} Our system has been evaluated on a vehicle platform with four fisheye cameras mounted for a parking assistance purpose, similar to \cite{schwesinger2016automated}. All cameras have a FoV of $190^{\circ}$ and output $1920\times1208$ color images at 25 frames per second (FPS). The vehicle equipped with a Nvidia Drive PX2 platform as the computation unit for autonomous driving. All cameras are connected to the PX2 through GMSL interface, while the vehicle status data such as the speed, steering wheel angle and the IMU measurements can be accessed through CAN bus with a rate of 30Hz. Since the testing scenarios include indoor environment, we use a fusion of vehicle odometry and high-accuracy IMU measurement as the ground truth. The on-line processing was run on PX2 to provide surround-view images, semantic results and vehicle odometry data. The off-line evaluation for VO was performed on a laptop with Intel Core i5 @ 2.3 GHz.

\textbf{Semantic segmentation model:} We choose the image cascade network (ICNet) \cite{zhao2018icnet} as the semantic segmentation model because of its excellent real-time performance with decent quality comparing with other high-accuracy models. A pre-trained ICNet model on Cityscapes \cite{cordts2016cityscapes} dataset is fine-tuned with our own labeled data. For a better segmentation performance, we respectively trained two models for outdoor scenario with 553 images and for indoor scenario with 360 images. We also used data augmentation techniques such as random resizing between 0.5 to 2 with a step of 0.5, horizontal flipping and adding color channels noise. For a further speed-up on PX2, a down-sampled $384\times384$ color image input was used as the model input which leads to an average processing rate of 7 FPS.

\textbf{Virtual LiDAR settings:} The generated free space image from segmentation model was processed with a morphology open operation with a kernel size of 2 to remove separate small obstacles. To remove too small obstacles in large free space areas, a minimum area threshold of 50 was set. The extracted contour points within a distance of 10 pixels to the border of the image were ignored. The angle increment of the simulated scan was set to $1^{\circ}$ for down-sampling the contour points. The input $384\times384$ image corresponds to a $15.3\:m\times15.3\:m$ area. Therefore the factor from pixel to meter is 0.03984 which provides an simple way to convert the pixel distance to practical distance for each measurement. The default parameters of the canonical scan matcher \cite{censi2008icp} were used.

\textbf{VO settings:} For each input surround-view image, about 500 ORB feature points \cite{rublee2011orb} were detected for matching. A maximum translation threshold for two corresponding feature points was set to $0.1\:m$ for the direct matcher. For the keyframe selection, thresholds of maximum translation of $1.5\:m$, rotation of $0.6\:rad$ and passed time of 3 seconds were chosen for the request of bundle adjustment. Since the pure feature based and pure scan based VO may fail to get a valid solution for relative pose estimation, we also checked the results of VO with the vehicle odometry. If the variation exceeded the translation and rotation thresholds of $0.2\:m$ and $0.1\:rad$, the output of VO was replaced with the transformation calculated by vehicle odometry.

Since both virtual LiDAR scans and visual features are provided, the proposed VO can work in three modes: \emph{scan-only} mode, \emph{feature-only} mode and \emph{scan+feature} mode. In \emph{scan-only} mode, the relative pose estimation is performed by the scan matcher, while the pose of keyframe is optimized with local BA. The \emph{feature-only} mode works in a similar way. In \emph{scan+feature} mode, the VO works the same as the pipeline illustrated in Fig.~\ref{scheme}. For the weights of feature matching and scan matching loss, they can be adjusted according to the amount and quality of feature and scan points. In our case we chose $1.0$ and $0.1$ for feature matching and scan matching loss, respectively. 

\begin{figure}[t]
	\centering
	\includegraphics[width=3in]{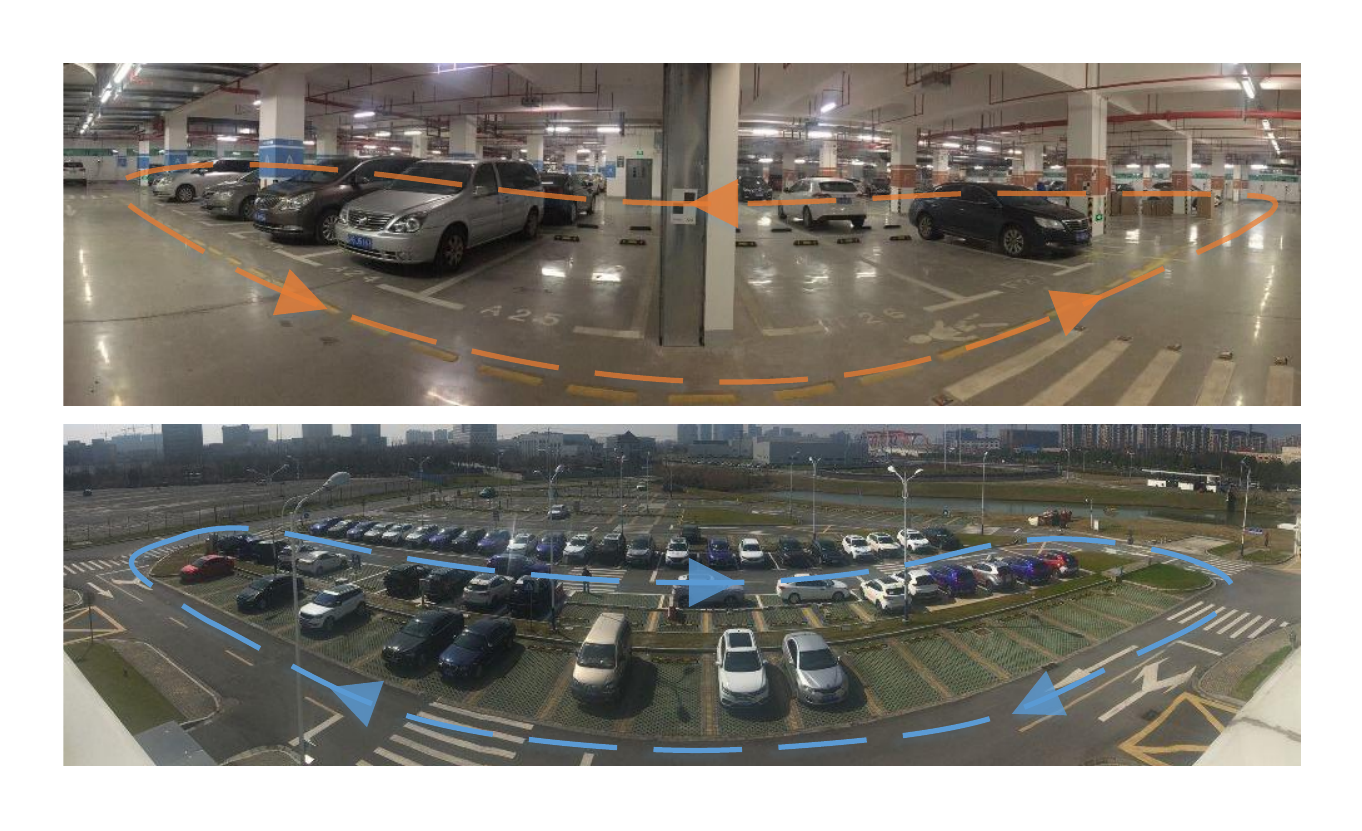}
	\caption{Photos of the testing scenario and navigation route: indoor garage (top) and outdoor parking lot (bottom).}
	\label{scenario}
\end{figure}

\begin{figure}[!tb]
\centering
\subfigure[Outdoor-park]{
\centering
\includegraphics[width=0.3\textwidth]{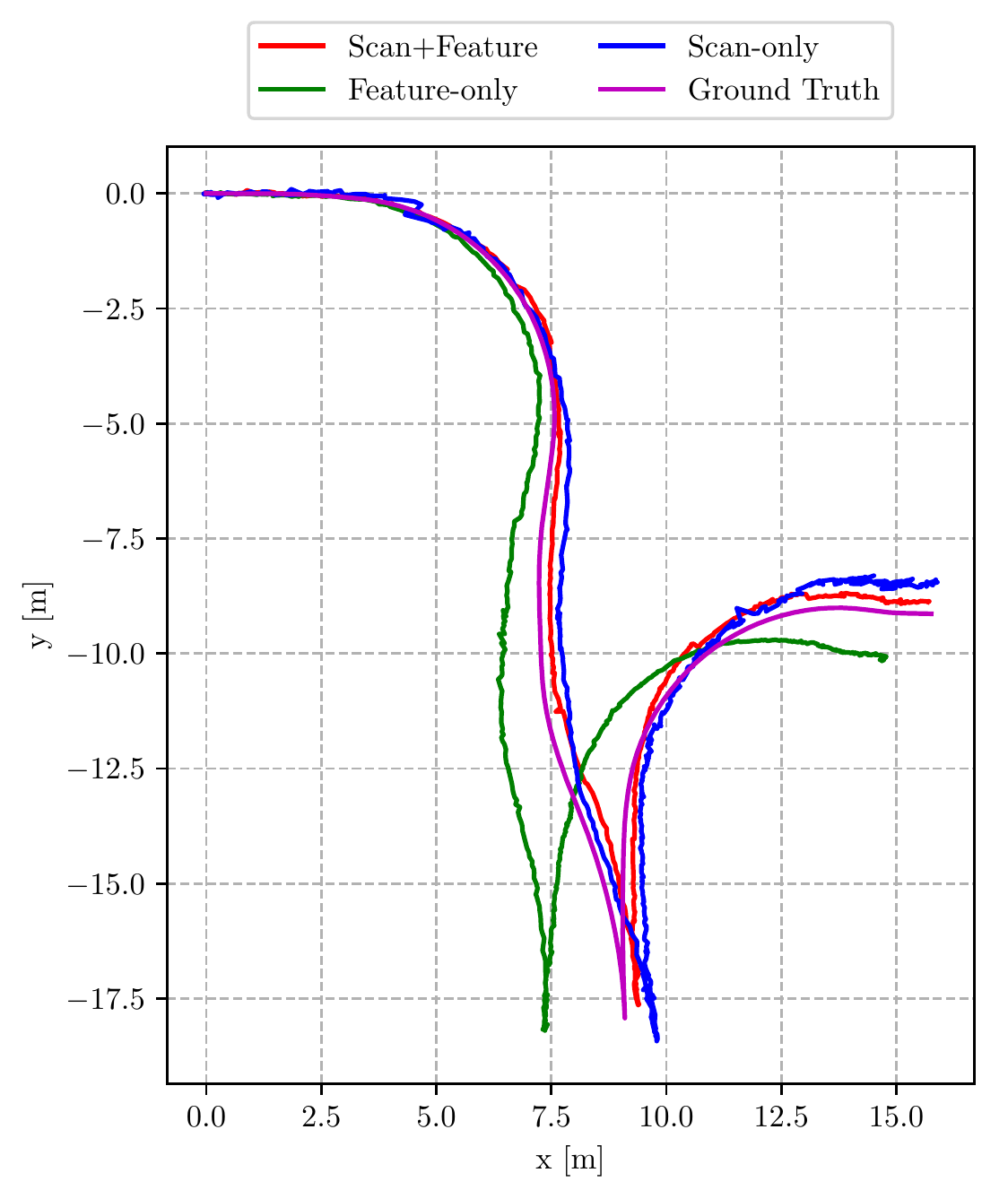}
\label{modes_outdoor_park}
}

\hspace{.1in}
\subfigure[Outdoor-navi]{
\centering
\includegraphics[width=0.47\textwidth]{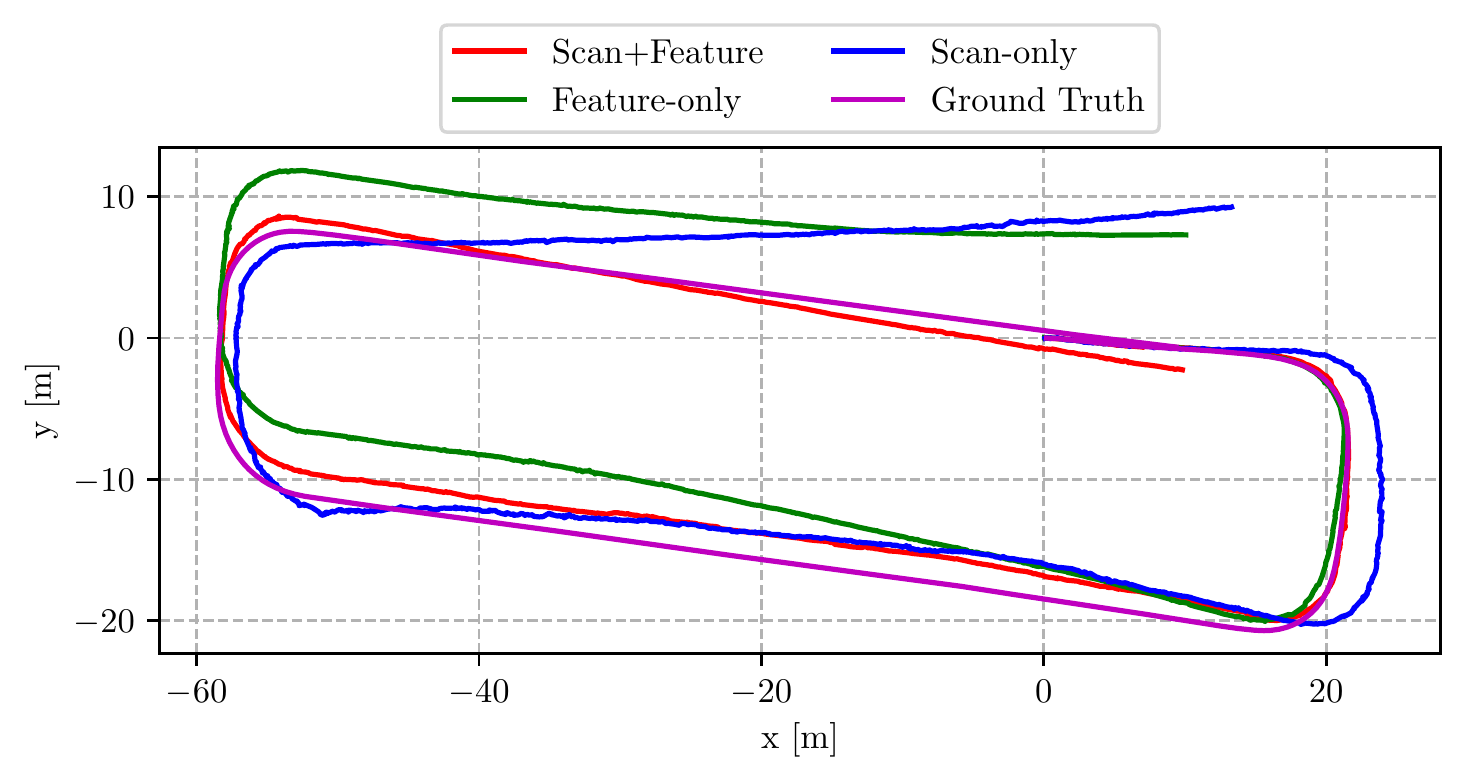}
\label{modes_outdoor_navi}
}

\vspace{0.01in}
\caption{Trajectories of ViLiVO in three modes in outdoor scenarios.}
\label{modes_outdoor}
\end{figure}

\begin{figure}[!htb]
\centering
\subfigure[Indoor-park]{
\centering
\includegraphics[width=0.47\textwidth]{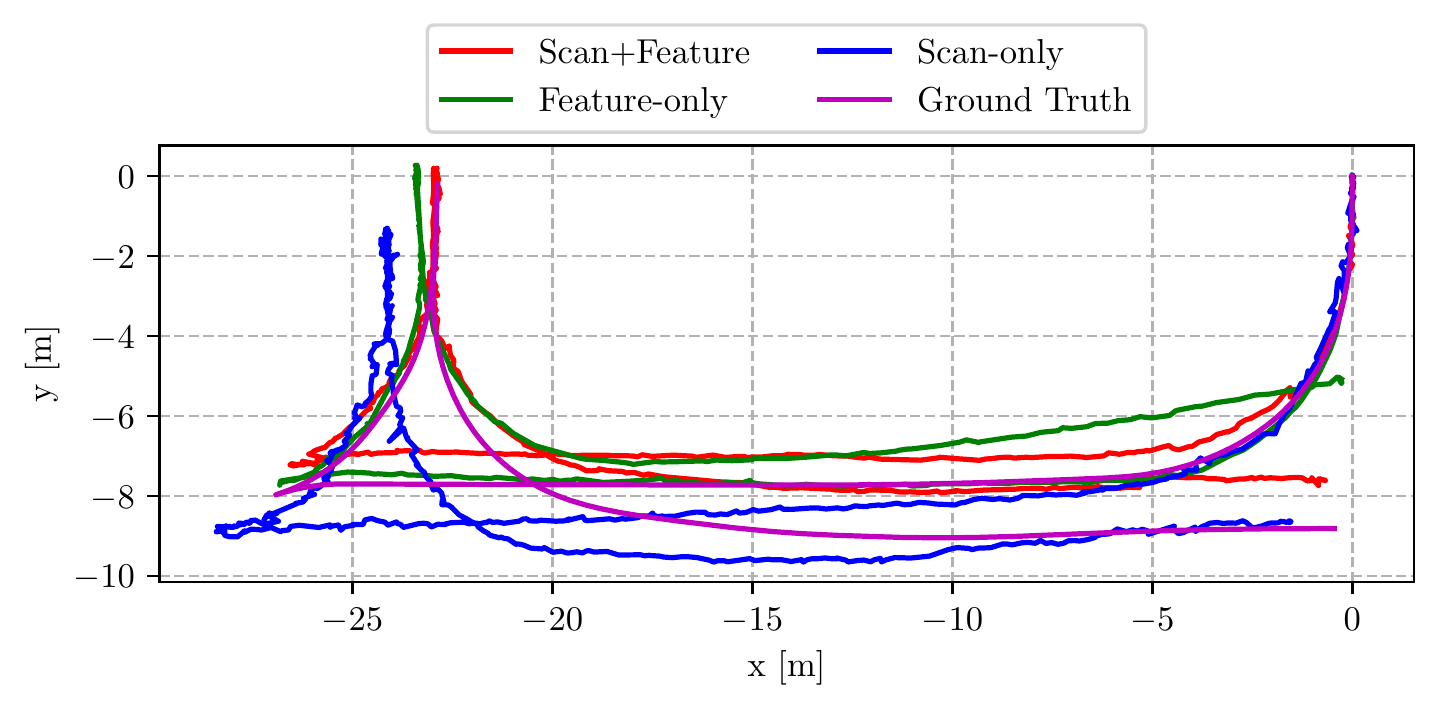}
\label{modes_indoor_park}
}

\hspace{.1in}
\subfigure[Indoor-navi]{
\centering
\includegraphics[width=0.47\textwidth]{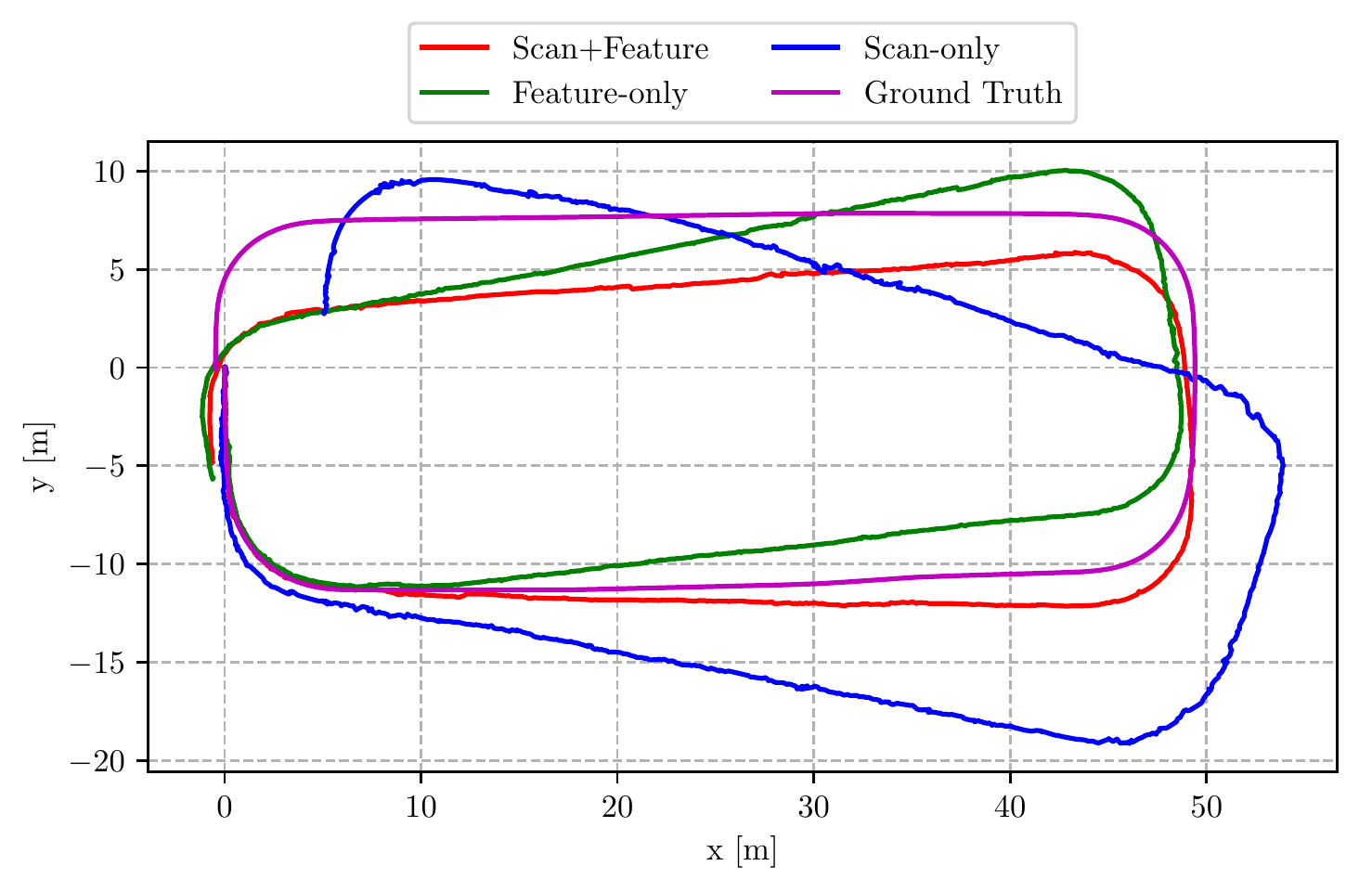}
\label{modes_indoor_navi}
}

\vspace{0.01in}
\caption{Trajectories of ViLiVO in three modes in indoor scenarios.}
\label{modes_indoor}
\end{figure}

\subsection{Testing Scenarios}
\textbf{Dataset:} Two indoor sequences in the garage and two outdoor sequences in the parking lot are selected for testing the proposed VO system.  Each two sequences contain one short distance and one long distance driving. The shorter one is used to simulate the parking process while the longer one is for navigating and searching an available parking space. The characteristics of four sequences are described in Table \ref{table1} and the photos of the scenarios are shown in Fig.~\ref{scenario}.

\begin{table}[!htbp]
\renewcommand{\arraystretch}{1.3}
\caption{Datasets for evaluation}
\label{table1}
\centering
\begin{tabular}{|c|c|c|c|}
\hline
\textbf{Name} & \makecell{\textbf{Length} \\ \textbf{(m)}}  & \makecell{\textbf{Mean Speed} \\ \textbf{(m/s)}} & \makecell{\textbf{Max. Speed} \\ \textbf{(m/s)}} \\
\hline
Outdoor-park &   37.12  &  0.30   &   0.71  \\
\hline
Outdoor-navi &   198.50  &  0.68   &  0.97   \\
\hline
Indoor-park &  69.90   &   0.52  &   1.15  \\
\hline
Indoor-navi &  128.38   &   0.64  &   0.96  \\
\hline
\end{tabular}
\end{table}


\textbf{Evaluation metrics:} A quantitative trajectory evaluation tool introduced in \cite{zhang2018tutorial} was used to compare the performance of different algorithms. The Relative Error (RE) results of trajectory segments with different length were calculated. Since the scale was undetermined for the baseline method, we applied the trajectory alignment before comparing our method with the baseline method.

\textbf{Baseline method:} We selected ORB-SLAM2 \cite{mur2017orb} as the baseline method. Since the fisheye camera is not originally supported by ORB-SLAM2, we built our own fisheye camera version instead of using undistorted images as input, which maximized the performance of the baseline system in our testing platform.
\subsection{Results and Discussions}
\begin{figure}[!t]
	\centering
	\includegraphics[width=3.3in]{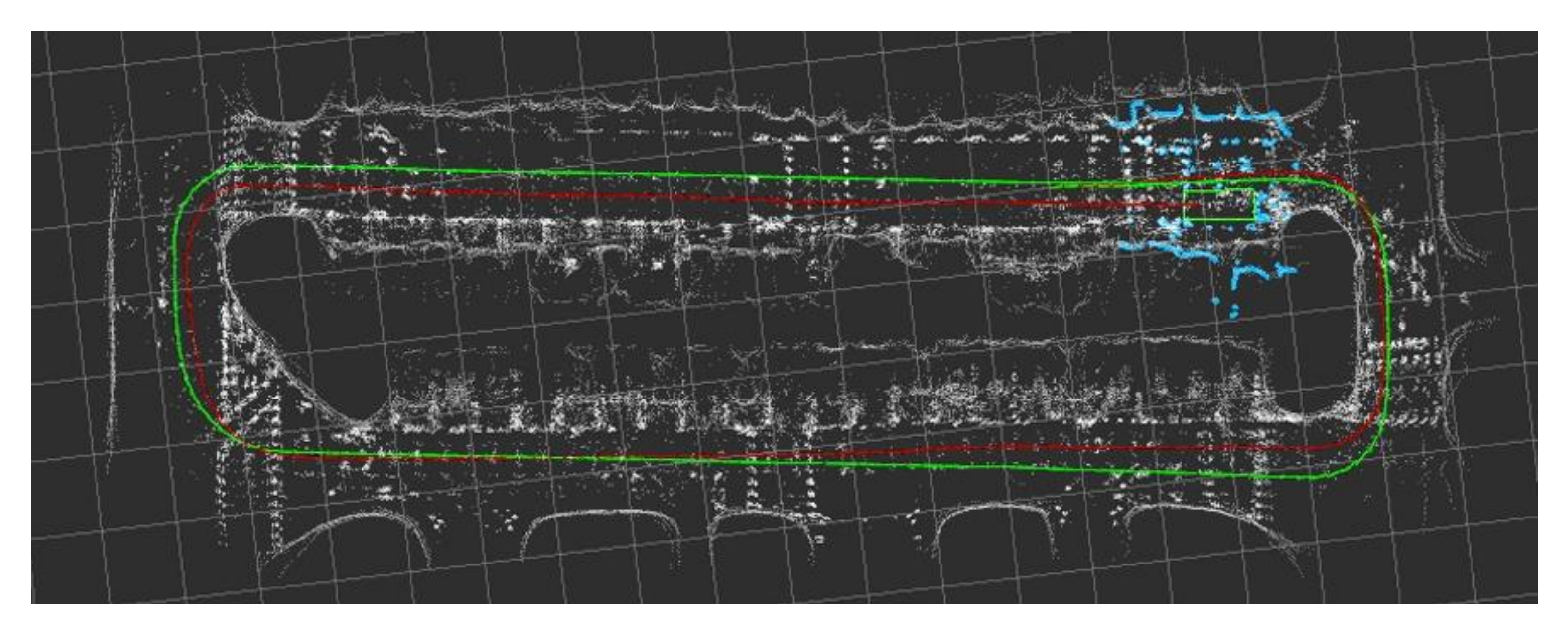}
	\caption{The output of ViLiVO in $\emph{scan+feature}$ mode in Outdoor-navi scenario, which includes the trajectory (red arrows), point cloud (white points), vehicle footprint (green rectangle), matched feature points in current keyframe (blue points). The ground truth is shown as green arrows. }
	\label{vilivo_output}
\end{figure}

\textbf{Different modes of ViLiVO:} We evaluated all three modes on all sequences for a comparison of accuracy. Fig.~\ref{modes_outdoor} and Fig.~\ref{modes_indoor} depicts the trajectories generated by ViLiVO in different modes. Fig.~\ref{vilivo_output} shows the typical output of ViLiVO in $\emph{scan+feature}$ mode.

\begin{figure}[!tb]
\centering
\subfigure[Translation errors in percentage]{
\centering
\includegraphics[width=0.44\textwidth]{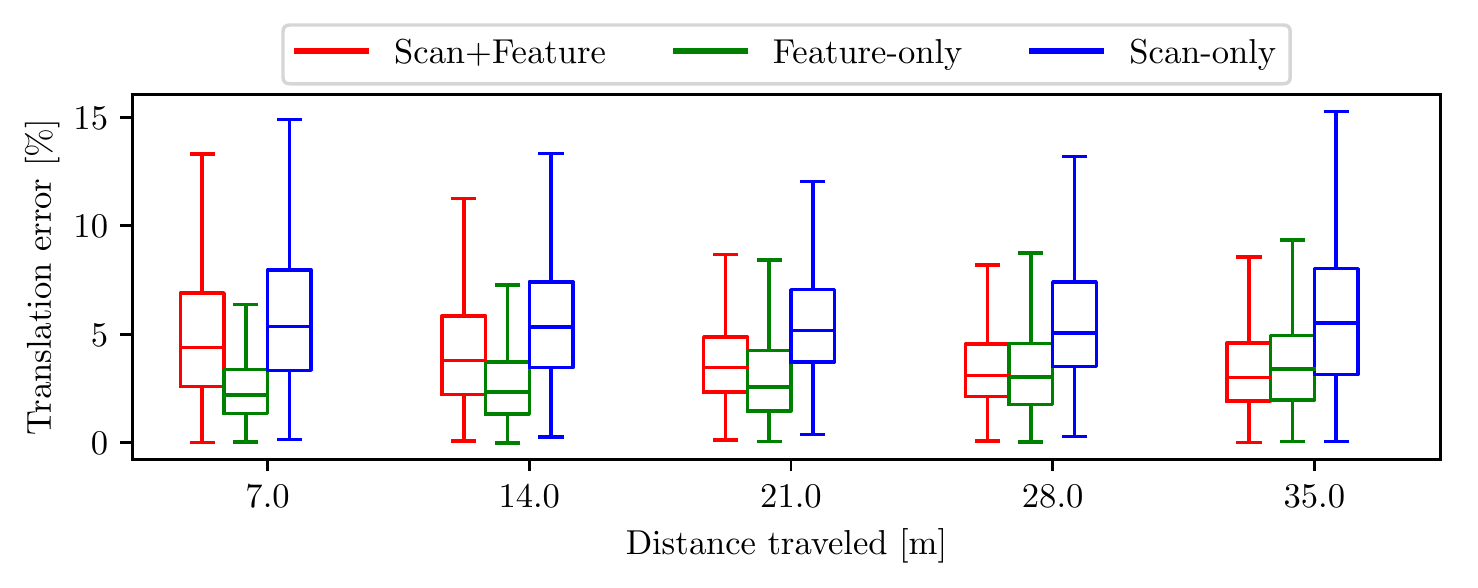}
\label{modes_relative_error_trans}
}

\hspace{.1in}
\subfigure[Rotation errors in degree]{
\centering
\includegraphics[width=0.44\textwidth]{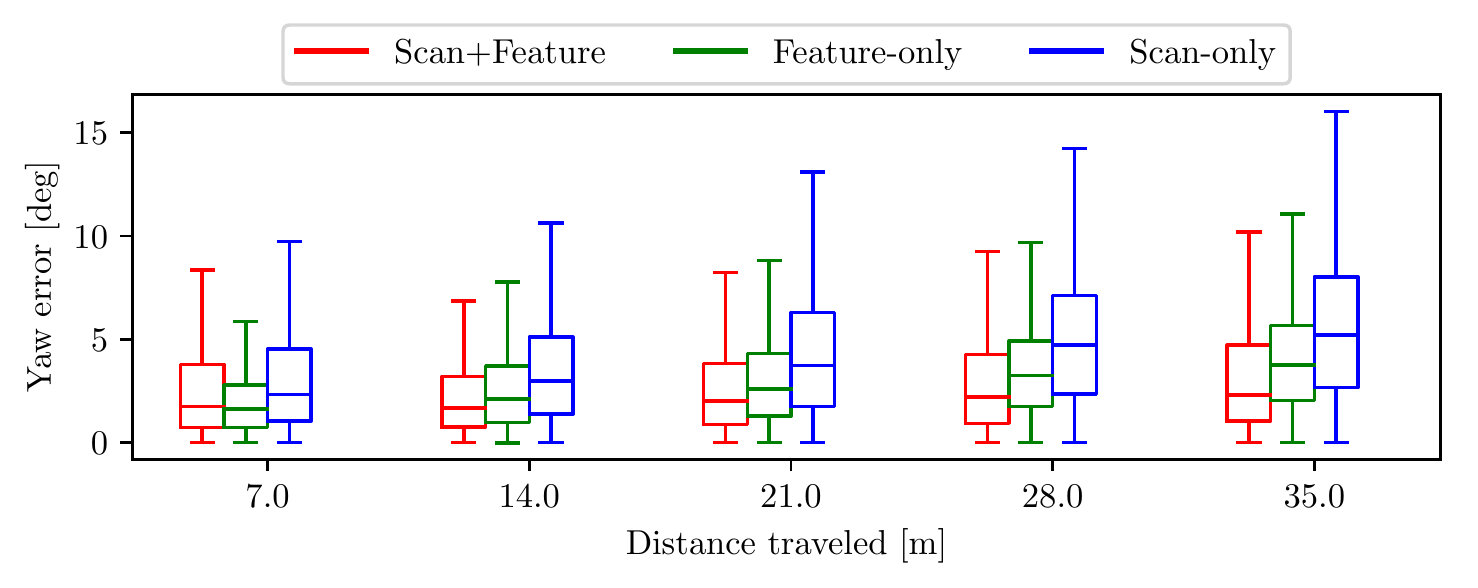}
\label{modes_relative_error_rot}
}

\vspace{0.01in}
\caption{Overall RE results for three modes of ViLiVO including both indoor and outdoor scenarios.}
\label{modes_relative_error}
\end{figure}

The calculated RE results are illustrated in Fig.~\ref{modes_relative_error}. From the translation error results we can find that the \emph{scan-only} mode has the worst performance with nearly two times more drift than the other two modes. This is mainly because the segmentation results are with significant noise and the number of matched scan points is usually a half of feature points. A more accurate segmentation model may further enhance the performance of the \emph{scan-only} mode. As for the $\emph{scan+feature}$ mode, its translation error is slightly over the \emph{feature-only} mode, which also reflects the influence of inaccurate scan measurements.

For the rotation error results, we can observe that the mode with both scan and feature information performances better than the \emph{scan-only} or \emph{feature-only} modes. It suggests that the rotation estimation can benefit from the combination of near feature points and far scan points.

\begin{figure}[t]
  \centering
  \includegraphics[width=3.4in]{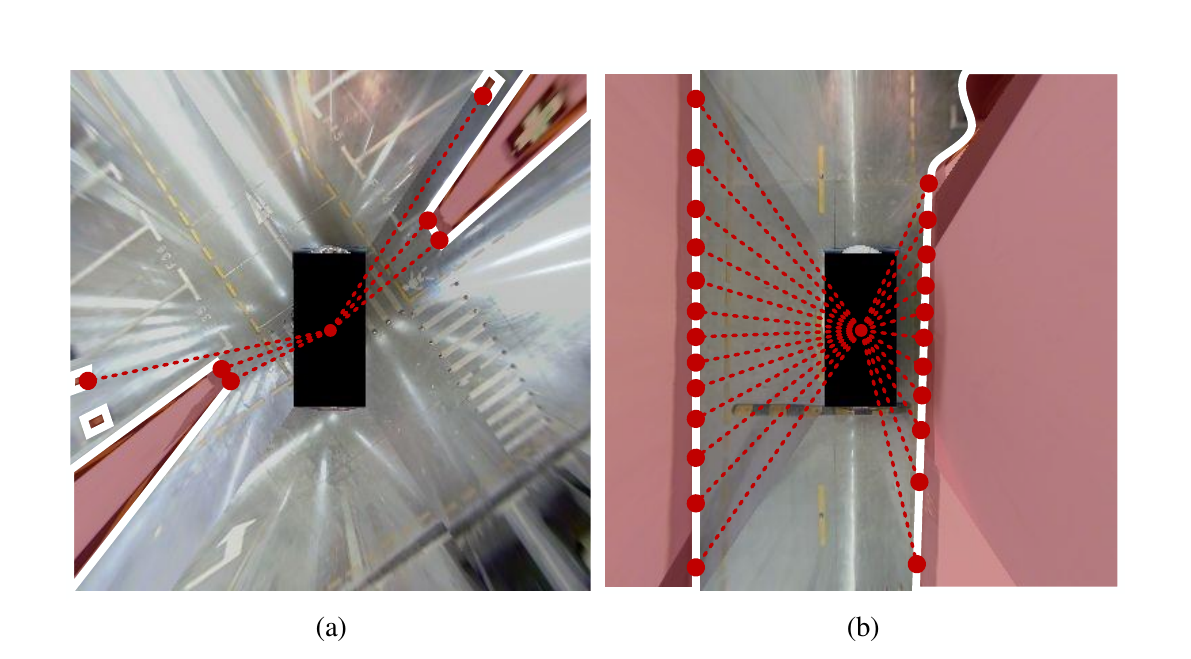}
  \caption{Demonstration of the complement between scan and feature points: few scan points with rich feature points (left) and few feature points with rich scan points (right).}
  \label{scan_feature_complement}
\end{figure}

For a comparison of robustness, in the experiments we have found that both \emph{scan-only} and \emph{feature-only} mode may failed to give a valid relative pose estimation under some difficult situations shown in Fig.~\ref{scan_feature_complement}. Obviously the scan measurements in Fig.~\ref{scan_feature_complement}(a) are too little for scan matching but with enough feature points for feature matching and vice versa for the situation in Fig.~\ref{scan_feature_complement}(b). Therefore the fusion of scan and feature points is a more robust and reliable input for the VO system.


\addtolength{\textheight}{-0.5cm}   

\begin{figure}[!t]
\centering
\subfigure[Outdoor-park]{
\centering
\includegraphics[width=0.47\textwidth]{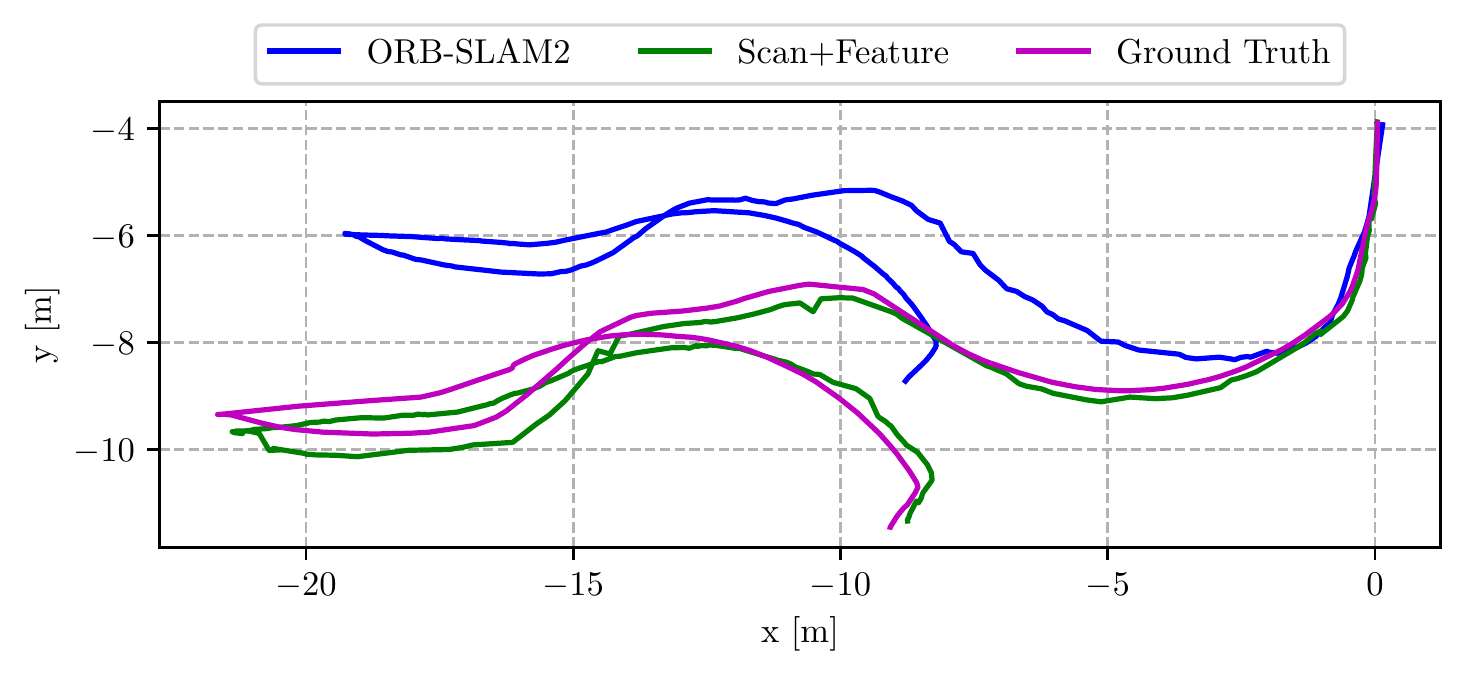}
\label{orb_outdoor_park}
}

\hspace{.1in}
\subfigure[Outdoor-navi]{
\centering
\includegraphics[width=0.47\textwidth]{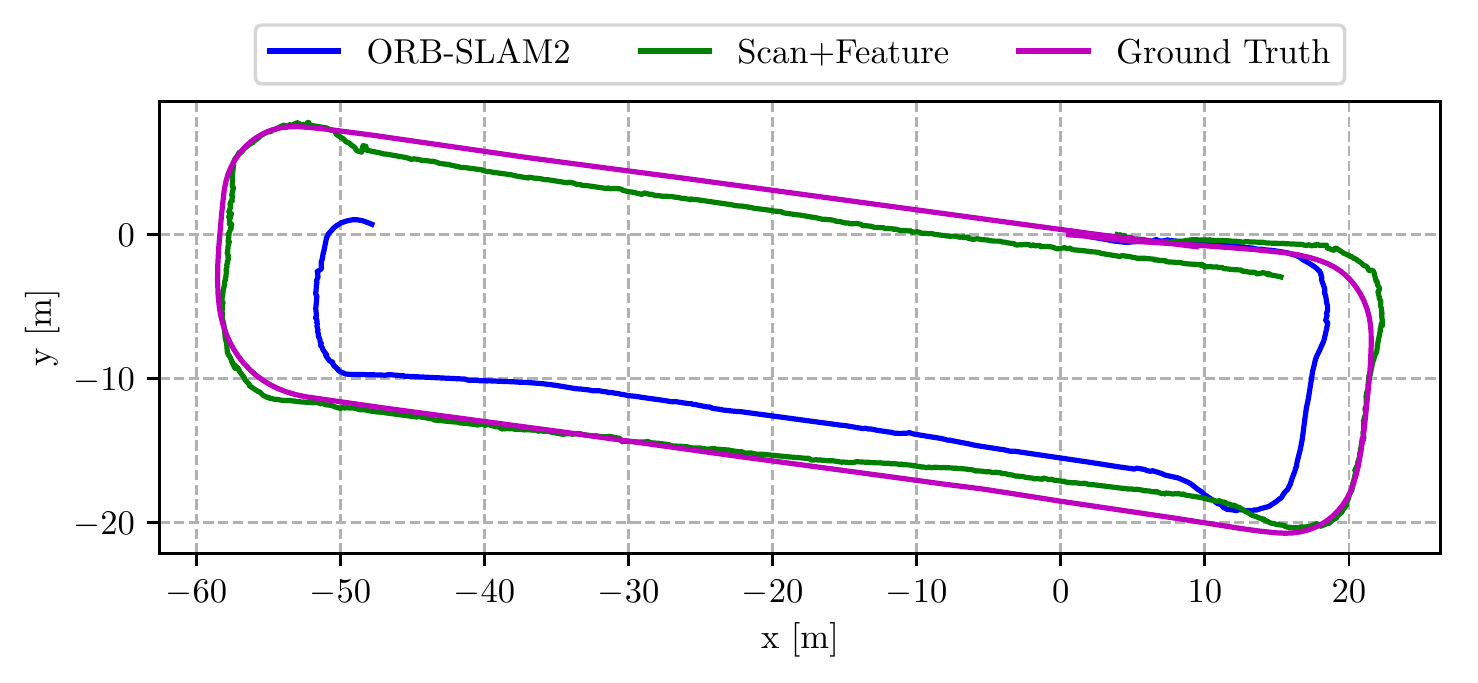}
\label{orb_outdoor_navi}
}

\vspace{0.01in}
\caption{Trajectories of ViLiVO ($\emph{scan+feature}$ mode) and ORB-SLAM2 in outdoor scenarios.}
\label{orb_outdoor}
\end{figure}

\begin{figure}[t]
\centering
\subfigure[Indoor-park]{
\centering
\includegraphics[width=0.47\textwidth]{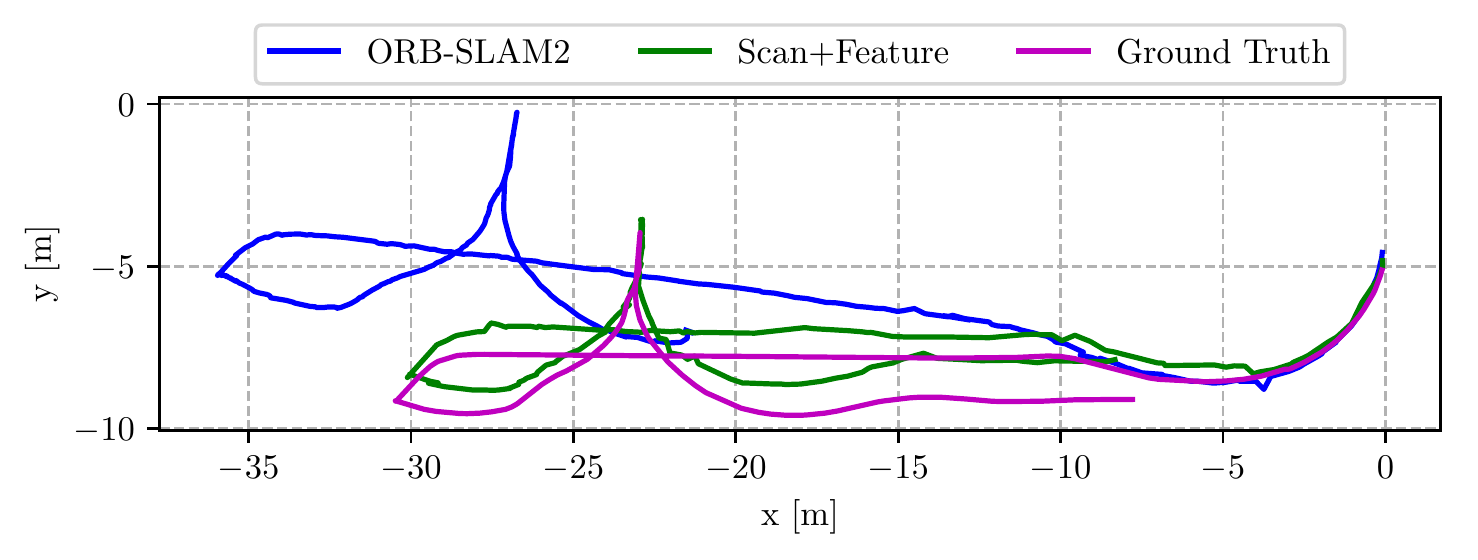}
\label{orb_indoor_park}
}

\hspace{.1in}
\subfigure[Indoor-navi]{
\centering
\includegraphics[width=0.47\textwidth]{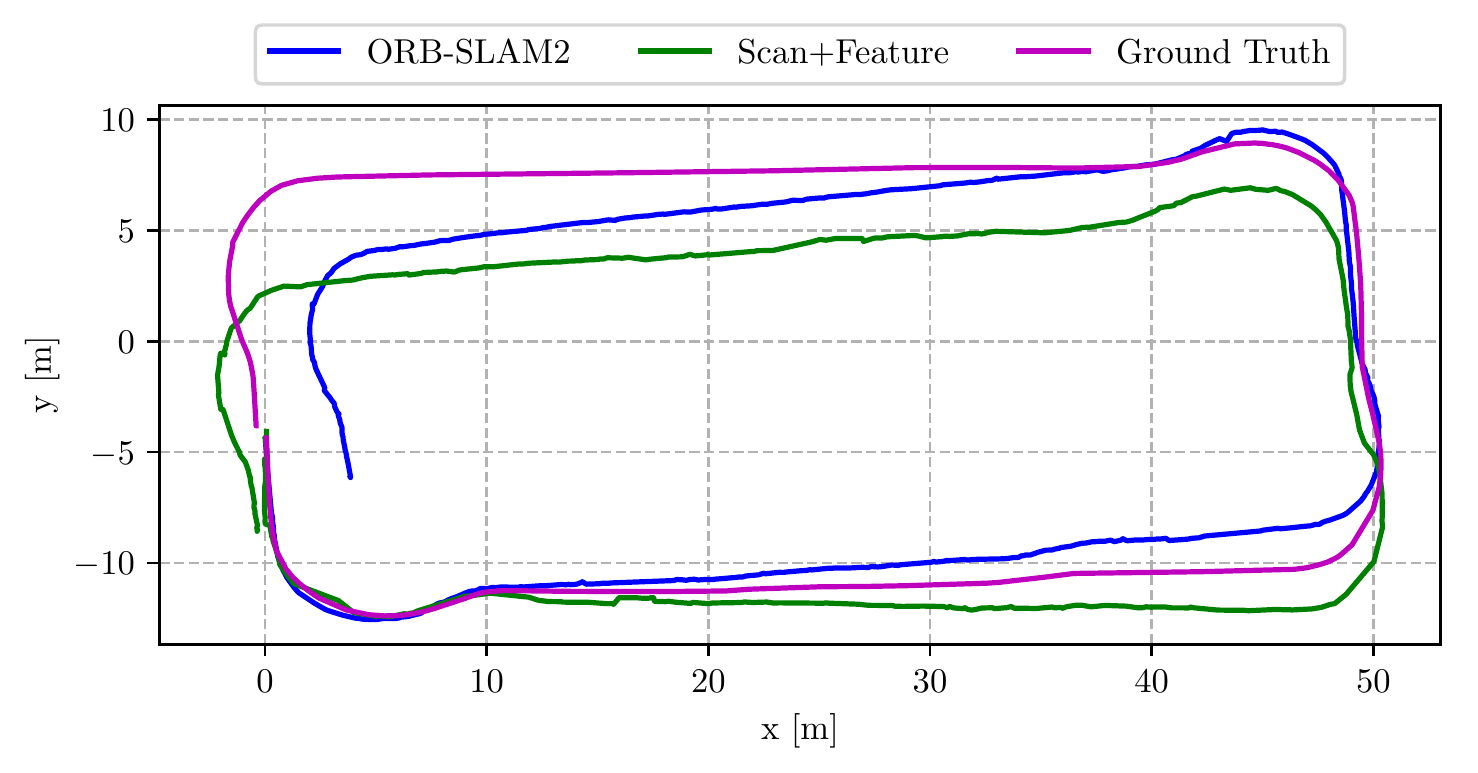}
\label{orb_indoor_navi}
}

\vspace{0.01in}
\caption{Trajectories of ViLiVO ($\emph{scan+feature}$ mode) and ORB-SLAM2 in indoor scenarios.}
\label{orb_indoor}
\end{figure}

\textbf{Comparing with baseline method:} The trajectories generated by ViLiVO in $\emph{scan+feature}$ mode and ORB-SLAM2 with the front fisheye camera are compared in Fig.~\ref{orb_outdoor} and Fig.~\ref{orb_indoor}. In the experiments we have found that ORB-SLAM2 failed to accomplish the mission frequently although using original fisheye images as input. For example, in Fig.~\ref{orb_outdoor}(b), ORB-SLAM2 lost tracking in the fourth turning. The failure situations of ORB-SLAM2 most occurred when turning at the corners or being too close to the wall and other vehicles. In such situations, many visual features are moving fast with motion blur, which makes it difficult for ORB-SLAM2 to keep stable tracking. For our system, on one hand, the virtual scans generated by semantic segmentation are robust to fast movement. On the other hand, the surround-view image captures more high-quality visual features than a single front view. Therefore our system has achieved a more robust performance.
 
\begin{figure}[!tb]
	\centering
	\subfigure[Translation errors in percentage]{
		\centering
		\includegraphics[width=0.44\textwidth]{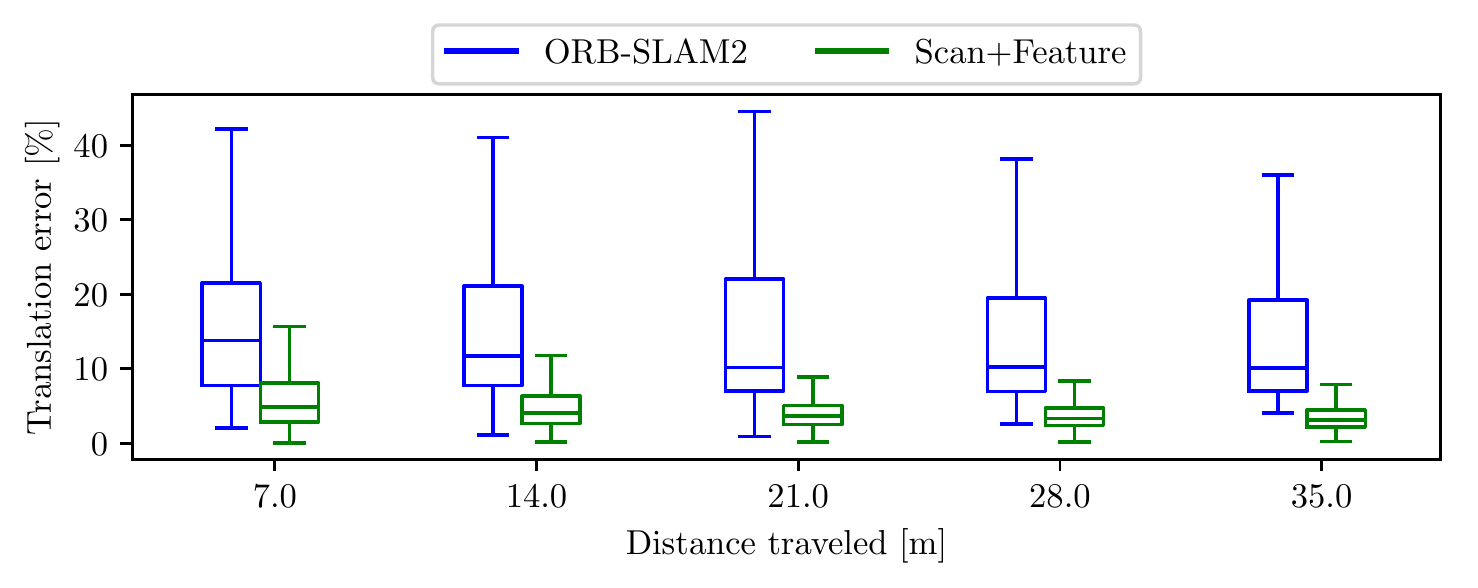}
		\label{orb_relative_error_trans}
	}
	
	\hspace{.1in}
	\subfigure[Rotation errors in degree]{
		\centering
		\includegraphics[width=0.44\textwidth]{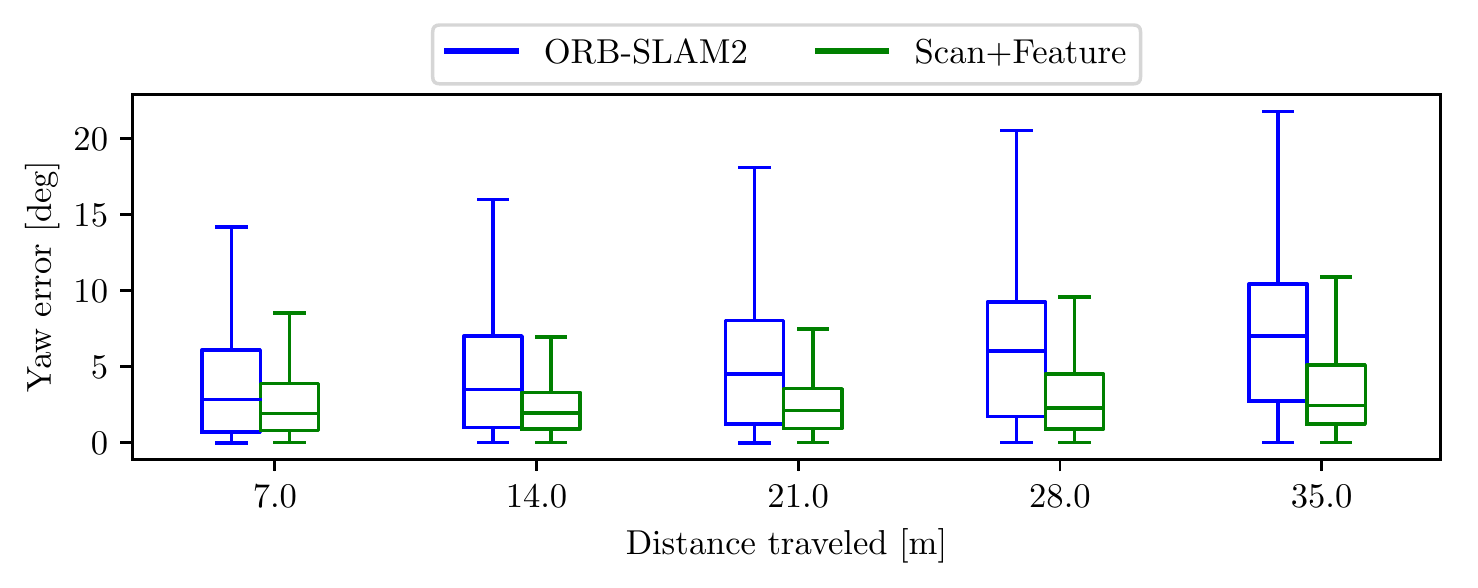}
		\label{orb_relative_error_rot}
	}
	
	\vspace{0.01in}
	\caption{Overall RE results for ViLiVO ($\emph{scan+feature}$ mode) and ORB-SLAM2 including both indoor and outdoor scenarios.}
	\label{orb_relative_error}
\end{figure}

The RE results are compared in Fig.~\ref{orb_relative_error}. Since the monocular ORB-SLAM2 is unable to output a practical scale for translation, the trajectories are first aligned with the ground truth with the initial 100 poses. However, we still found an obvious scale drift in the experiments of ORB-SLAM2 which resulted in large translation error. Comparatively speaking, our method remains a stable performance among all sequences.

\section{Conclusion}
In this paper we integrate LiDAR scan information into a VO system. Instead of using a physical LiDAR, we apply a semantic segmentation model to process the surround-view image around the vehicle to work as a virtual LiDAR. The scan data of our virtual LiDAR can be extracted from the contours of the free space area. As for the motion estimation, our ViLiVO system merges both the feature matching and the scan matching information in the BA optimization procedure to achieve a better pose estimation for the keyframe. Comparing with the pure feature-based VO, experimental results show that the proposed VO system leverages both the texture and geometric information which leads to a more robust performance. In addition to working standalone, our VO system can combine with a loop closure and global BA module to form a complete VSLAM system, and furthermore provide obstacle information to construct a visual navigation system.




\bibliographystyle{IEEEtran}
\bibliography{ref}

\end{document}